\documentclass{article}
  \usepackage[main, preprint]{neurips_2025}

\usepackage[utf8]{inputenc} % allow utf-8 input
\usepackage[T1]{fontenc}    % use 8-bit T1 fonts
\usepackage{hyperref}       % hyperlinks
\usepackage{url}            % simple URL typesetting
\usepackage{booktabs}       % professional-quality tables
\usepackage{amsfonts}       % blackboard math symbols
\usepackage{nicefrac}       % compact symbols for 1/2, etc.
\usepackage{microtype}      % microtypography
\usepackage{xcolor}         % colors
\usepackage{enumitem}
\usepackage{amsmath}
\usepackage{subcaption}
\usepackage{graphicx}
\usepackage{multirow}
\usepackage{threeparttable}

\title{Beyond Compression: Quantifying Spectral Accessibility in Vision Representations}

\author{%
  Akayou A.~Kitessa \\
  Fordham University \\
  New York City, NY \\
  \texttt{ak214@fordham.edu} \\
  \And
  Yijun Zhao \\
  Fordham University \\
  New York City, NY \\
  \texttt{yzhao11@fordham.edu} \\
}

\begin{document}

\maketitle

\begin{abstract}

Vision-language models map visual features into a shared embedding space through learned projection layers, yet it remains unclear how these transformations alter the structure of visual information. This study examines changes in representation through spatial-frequency accessibility, measured by the linear recoverability of band-limited Fourier energy from model representations. To isolate effects beyond dimensionality reduction, we introduce Residual Spectral Loss (RSL), which evaluates changes relative to a dimension-matched random projection baseline. To reduce confounding effects from optimization, the analysis uses pretrained models with all parameters frozen. The experimental results show consistent frequency-dependent changes in accessibility across CLIP and DINOv2 on ImageNet and MS-COCO datasets. Spectral accessibility follows a non-monotonic trajectory across depth, peaking at intermediate layers before decreasing toward the output representation. The final transformation differs across architectures: CLIP’s learned projection is spectrally neutral, with changes explained by compression, whereas DINOv2’s \texttt{[CLS]} pooling induces a structured loss across the spectrum. These findings identify intermediate layers and pooling mechanisms as primary drivers of spectral transformation in modern vision encoders.

\end{abstract}

\section{Introduction}
\label{sec:intro}

Vision-language models such as CLIP \citep{radford2021learning} map high-dimensional visual features into a shared multimodal embedding space through a learned projection layer. This component, referred to here as the \emph{connector}, is widely used in modern architectures. However, relatively little is known about how this mapping alters the fundamental structure of the visual information it carries. A common assumption is that such projections act as a neutral form of dimensionality reduction and discard information in an unstructured manner. In contrast, these connectors are learned under objectives that link visual features with text and may systematically bias the representation toward features that support this coupling. This motivates the central question: \emph{does the representation selectively reshape spectral content, and does this deviation exceed what would be expected from compression alone?}

We study this question through spatial-frequency accessibility. Instead of reconstructing pixels or intermediate features, the analysis tests whether specific frequency components of the original image are recoverable from the representation through a linear mapping. This framework provides a controlled setting to quantify how representations preserve or suppress information across the frequency spectrum and builds on prior work that uses linear probes to assess representational accessibility \citep{alain2016understanding}. It also relates to evidence that deep visual models exhibit systematic frequency biases, including a preference for texture over shape and sensitivity to high-frequency components \citep{geirhos2019imagenet,wang2020high}. Because the analysis relies on linear decoding, it quantifies the accessibility of frequency components rather than the total information content of the representation.

We introduce Residual Spectral Loss (RSL) to quantify additional changes in accessibility induced by learned transformations relative to a dimension-matched random projection baseline that captures the effect of compression alone. Experiments are conducted on subsets of the ImageNet dataset \citep{deng2009imagenet} and the COCO dataset \citep{lin2014coco}. The vision encoders are pretrained and kept frozen. Only the linear probes are fit. This approach limits confounding effects from model fine-tuning.

The analysis is evaluated across multiple architectures, including CLIP ViT-L/14, CLIP ViT-B/32, and DINOv2 ViT-B \citep{oquab2023dinov2}. Multiple representational stages are investigated, including the convolutional stem \citep{Xiao2021Early}, intermediate layers, and the final projection. Results show statistically significant, frequency-dependent changes in accessibility, with the largest losses consistently occurring at intermediate layers. While the projection stage also introduces structured changes, the dominant effect appears prior to the final mapping. These findings indicate that learned transformations induce a distinct spectral signature that cannot be explained by dimensionality reduction alone.

\paragraph{Contributions.}

\begin{enumerate}[leftmargin=2.5em, itemsep=0pt]
    \item A framework for measuring spectral accessibility. Linear probes quantify the recoverability of radial frequency components and characterize preserved information.

    \item A method to isolate learned transformation effects from dimensionality reduction. A dimension-matched random baseline and RSL separate compression effects from changes induced by learned mappings.

    \item Empirical evidence of consistent, frequency-dependent accessibility changes across CLIP and DINOv2 models. These results demonstrate that training objectives influence spectral properties of representations.

    \item A layerwise analysis demonstrating that the largest accessibility losses occur at intermediate layers, rather than uniformly across the network.
\end{enumerate}

\section{Related Work}
\label{sec:related_work}

\subsection{Probing Representation Accessibility in Deep Models}

A common approach to analyzing representations in deep networks is to evaluate what information remains accessible through simple downstream mappings. Linear probes provide a controlled way to assess whether a representation retains information relevant to a target variable without requiring end-to-end retraining \citep{alain2016understanding}. In this framework, probe performance reflects the degree to which information is linearly accessible from a representation. This approach is well suited to layerwise analysis, where the goal is to characterize how learned transformations preserve or suppress information across depth. In this work, we apply linear probing in the frequency domain by predicting band-limited Fourier energy of the input image from intermediate and final representations, thereby extending probing from semantic attributes to spatial-frequency accessibility.

\subsection{Frequency Bias and Spectral Structure in Vision Models}

A growing body of work has shown that deep vision models exhibit systematic biases in how they encode spatial-frequency information. Prior work has shown that standard image classifiers often rely heavily on texture cues rather than global shape and can be sensitive to high-frequency patterns that are less salient to human observers \citep{geirhos2019imagenet, wang2020high}. These findings suggest that visual representations may emphasize particular frequency-dependent signals over others.

The frequency domain has also been used to analyze architectural behavior more directly. Prior work has argued that self-attention in Vision Transformers (ViTs)~\citep{dosovitskiy2021image} behaves as a low-pass operator, progressively attenuating higher-frequency components with depth \citep{wang2022antioversmoothing}. Other studies evaluate robustness through Fourier-based perturbations, showing that Fourier analysis can expose systematic vulnerabilities and spectral preferences in computer vision systems \citep{yin2019fourier}. Frequency-based analyses have additionally been applied to generative models, where spectral discrepancies reveal artifacts and inductive biases that are not always apparent in pixel space \citep{schwarz2021frequency}.

Our work builds on this literature by focusing not only on whether models exhibit frequency bias, but on where frequency-dependent accessibility changes emerge across layers and whether those changes exceed what would be expected from dimensionality reduction alone.

\subsection{Information Flow in Vision Transformers}

Prior work has shown that ViTs develop internal representations that differ in important ways from those of convolutional networks. Comparative analyses suggest that ViTs exhibit relatively uniform representational structure across depth and incorporate global information at early layers, while residual connections help preserve lower-level spatial information \citep{raghu2021vision}. These findings motivate layerwise analyses aimed at identifying where information is retained or transformed.

Related work has also shown ViTs can remain robust under strong spatial perturbations, suggesting that meaningful spatial structure can remain accessible even as representations become increasingly global \citep{naseer2021intriguing}. At the same time, local information is not distributed uniformly across tokens. Large ViTs can develop high-norm tokens that act as global information aggregators while becoming less faithful carriers of local spatial and positional content \citep{darcet2024registers}.

These observations are directly relevant to spatial-frequency accessibility. Changes across layers may reflect not only overall compression, but also redistribution of information across token types and the effects of final aggregation.

\subsection{Training Objectives and Spectral Accessibility}
Our experiments compare models trained under different objectives, including multimodal contrastive learning in CLIP and self-supervised visual learning in DINOv2 \citep{radford2021learning,oquab2023dinov2}. Prior work has shown that training objectives strongly influence the learned visual representations, including their invariances, transfer properties, and internal organization. This makes objective-level comparison important when studying spectral accessibility: differences between CLIP and DINOv2 may reflect not only architectural similarities, but also the distinct pressures imposed by language alignment versus self-distillation.

In this context, our analysis complements existing comparisons of pretrained visual representations by examining how different training objectives affect accessibility across the spatial-frequency spectrum and across model depth. Rather than evaluating only downstream task performance, we study how objective choice shapes the internal preservation and suppression of visual information.

\section{Data and Preprocessing}
\label{sec:data}

We use subsets of the open-access ImageNet \citep{deng2009imagenet} and the COCO \citep{lin2014coco} datasets. For each dataset, a total of $N=10{,}000$ images are randomly sampled and resized to the input resolution required by each model (e.g., $224 \times 224$). The same subsets are used across all models to maintain comparability of extracted representations. Each dataset is partitioned into training and test sets using a fixed random seed, with an 80/20 split. The training set is used to fit all linear probes, while the test set is reserved exclusively for evaluation.

\subsection{Preprocessing and spectral representation.}
For each image $x \in \mathbb{R}^{H \times W}$, we first convert it to grayscale to isolate spatial structure. The image is then multiplied element-wise by a separable two-dimensional Hann window $W$:
\begin{equation}
x_w(i,j) = W(i,j)\, x(i,j), \qquad W(i,j) = w_H(i)\, w_W(j),
\end{equation}
where $w_H$ and $w_W$ are standard one-dimensional Hann windows \citep{harris1978windows}. The centered two-dimensional discrete Fourier transform is computed as
\begin{equation}
F = \mathrm{fftshift}(\mathcal{F}{x_w}),
\end{equation}
with power spectrum $P(i,j) = |F(i,j)|^2$. The zero-frequency (DC) component is removed to avoid dominance by global intensity.

\subsection{Frequency band construction}

Natural images exhibit an approximate $1/f$ power spectrum \citep{torralba2003statistics, vanderSchaaf1996modelling}, so we partition the frequency plane into $K=12$ logarithmically spaced radial bands to maintain adequate signal at higher frequencies. Let $R(i,j)$ denote the radial frequency at coordinate $(i,j)$, and let $(c_y, c_x)$ denote the center of the spectrum.

We define $K+1$ radial boundaries $\{r_k\}_{k=0}^K$ using logarithmic spacing:
\begin{equation}
r_k = \exp\left(\log r_{\min} + \frac{k}{K} (\log R_{\max} - \log r_{\min})\right), \quad k = 0, \dots, K,
\end{equation}
where $r_{\min} = 1$ excludes the DC component from the band construction and $R_{\max} = \max_{i,j} R(i,j)$.

These boundaries induce $K$ frequency bands:
\begin{equation}
\Omega_k = \{(i,j) \mid r_{k-1} \le R(i,j) < r_k\}, \quad k = 1, \dots, K.
\end{equation}

For each band $\Omega_k$, we compute the normalized energy
\begin{equation}
E_k(x) =
\frac{\sum_{(i,j)\in\Omega_k,\ (i,j)\neq(c_y,c_x)} P(i,j)}
{\sum_{(i,j)\neq(c_y,c_x)} P(i,j)},
\end{equation}
which represents the fraction of total non-DC Fourier energy contained in band $k$. These band energies $E_k(x)$ serve as targets in the spectral accessibility analysis (Section~4.2).

\begin{figure}[!t]
  \centering
\includegraphics[ trim = 305 100 220 65, clip, scale=0.32]{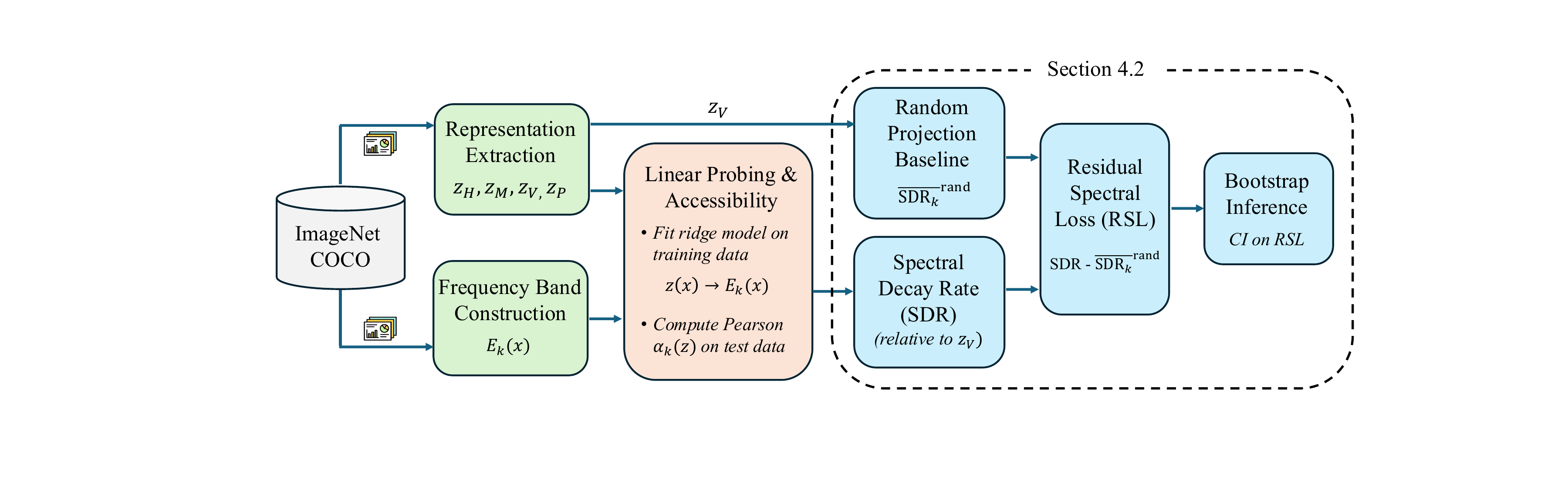}
 % \vspace{-1mm}
 \caption{ Pipeline for Spectral Accessibility Analysis
}
\label{fig:pipeline}
 \vspace{-1mm}
\end{figure}

% \begin{figure}[!t]
%   \centering
%   \includegraphics[width=\textwidth]{results/pipeline_diagram.png}
%   \caption{Overview of the experimental pipeline. Stage~1 extracts four representational stages from the models. Spatial-frequency targets are constructed via logarithmic radial partitioning of the 2D power spectrum. Stage~2 measures accessibility via linear probing, computes the Spectral Decay Rate (SDR) and Residual Spectral Loss (RSL) against a semi-orthogonal random projection baseline, and estimates confidence intervals via bootstrap resampling.}
%   \label{fig:pipeline}
% \end{figure}

\section{Methods}
\label{sec:methods}

Figure \ref{fig:pipeline} summarizes the analysis pipeline. Given an input image, we extract layer-wise representations from pretrained vision encoders and compute band-limited spectral energies as targets. Linear probes are then used to measure how well each representation predicts these frequency components. To separate learned effects from dimensionality reduction, we compare against a dimension-matched random projection baseline and quantify the difference using Residual Spectral Loss (RSL). The following subsections describe representation extraction (Section \ref{subsec:feature_extraction}) and spectral accessibility analysis (Section \ref{subsec:analysis}). Code to reproduce the full analysis pipeline is available at an anonymized GitHub repository \citep{github}. All experiments were conducted on a Linux system with NVIDIA V100 GPUs. The analysis can be reproduced on standard hardware.
%The experiment was performed on a Linux machine (Fedora, kernel 6.6.3) with an Intel(R) Xeon(R) Bronze 3104 CPU, approximately 157 GB RAM, and two NVIDIA GPUs: Tesla V100-PCIE-32GB and Tesla V100S-PCIE-32GB. However, since the computations involved are not resource-intensive, they should be performed on computers with fewer resources.

% ======================================================================
\subsection{Representation Extraction}
\label{subsec:feature_extraction}
% ======================================================================
We extract representations from pretrained vision encoders at multiple depths to characterize how visual information evolves across layers. This section specifies the models under study and the layer-wise representations used in the spectral accessibility analysis. All representations are computed with frozen model weights.

\subsubsection{Models}
\label{subsubsec:model}

We evaluate ViT~\citep{dosovitskiy2021image} architectures that differ in their training objective and in whether a language-aligned projection is present. All weights are frozen without fine-tuning.

\paragraph{CLIP ViT-L/14.}
CLIP~\citep{radford2021learning} is a contrastive vision-language model whose visual encoder is a ViT-Large with patch size 14, comprising $L{=}24$ transformer blocks and a learned linear projection mapping the final visual representation ($D{=}1024$) into a shared embedding space ($d{=}768$). We load the model via the OpenCLIP library~\citep{ilharco2021openclip} using the publicly released OpenAI checkpoint.

\paragraph{CLIP ViT-B/32.}
We also evaluate CLIP with a ViT-Base architecture and patch size 32. The encoder comprises $L{=}12$ transformer blocks with hidden dimension $D{=}768$, followed by a learned projection to a $d{=}512$-dimensional embedding space. This model serves as a comparison to assess whether spectral effects persist across model scale and patch resolution.

\paragraph{DINOv2.}
DINOv2~\citep{oquab2023dinov2} is a self-supervised vision model trained with self-distillation and masked image modeling, without language supervision. We use the ViT-Base variant (\texttt{facebook/dinov2-base}) loaded via the HuggingFace Transformers library~\citep{wolf2020transformers}. The encoder comprises $L{=}12$ transformer blocks with hidden dimension $D{=}768$. Unlike CLIP, DINOv2 has no projection to a language-aligned space.

\begin{table}[!t]
  \caption{Summary of representational stages extracted from each model.}
  \label{tab:feature_summary}
  \centering
  \small
  \setlength{\tabcolsep}{10pt}
  \begin{tabular}{llll}
    \toprule
    \textbf{Stage} & \textbf{Symbol} & \textbf{CLIP (ViT-B/32, ViT-L/14)} & \textbf{DINOv2} \\
    \midrule
    Conv. Stem & $\mathbf{z}_H$
      & Spatial avg.\ of \texttt{conv1} tokens
      & Spatial avg.\ of \texttt{conv} tokens \\
    Mid-layer & $\mathbf{z}_M$
      & \texttt{[CLS]} at block 11
      & Mean-pool patches at layer 6 \\
   Pre-projection  & $\mathbf{z}_V$
      & \texttt{[CLS]} at block 24
      & Mean-pool patches at layer 12 \\
    Output & $\mathbf{z}_P$
      & $\mathbf{z}_V W_{\text{proj}}$
      & \texttt{[CLS]} at layer 12 \\
    \bottomrule
  \end{tabular}
  \vspace{-2mm}
\end{table}

\subsubsection{Layer-wise Representations}
\label{subsubsec:representational_stages}

For each input image $\mathbf{x}$, we extract representations at four stages to characterize the evolution of spectral information across the encoder. Following \citet{Xiao2021Early}, we treat the initial projection as a \emph{convolutional stem}. CLIP and DINOv2 differ in their pooling mechanisms. CLIP relies on \texttt{[CLS]}, while DINOv2 uses patch-average pooling at intermediate layers and \texttt{[CLS]} at the output. Table~\ref{tab:feature_summary} summarizes the extraction points.

Let $\mathbf{t}_i^{(\ell)} \in \mathbb{R}^D$ denote the $i$-th token at layer $\ell$, and let $P$ denote the number of patch tokens. The representational stages are defined as follows:
\paragraph{Convolutional Stem ($\mathbf{z}_H$)}
$\displaystyle \mathbf{z}_H = \frac{1}{P} \sum_{i=1}^{P} \mathbf{t}_i^{(0)}$.
\paragraph{Mid-layer and Pre-projection ($\mathbf{z}_M$, $\mathbf{z}_V$)}
$\displaystyle \mathbf{z}^{(\ell)} = \begin{cases} 
\mathbf{t}_{\texttt{[CLS]}}^{(\ell)} & \text{CLIP}  \\[5pt]
\frac{1}{P} \sum_{i=1}^{P} \mathbf{t}_i^{(\ell)} & \text{DINOv2}
\end{cases}$ 

\hspace{6cm}where $\ell \in \{\lfloor L/2 \rfloor, L\}$, $\mathbf{z}_M = \mathbf{z}^{(\lfloor L/2 \rfloor)}$, and $\mathbf{z}_V = \mathbf{z}^{(L)}$.

\paragraph{Output Representation ($\mathbf{z}_P$)}
$\displaystyle \mathbf{z}_P = \begin{cases} 
\mathbf{z}_V W_{\text{proj}} & \text{CLIP} \\[5pt]
\mathbf{t}_{\texttt{[CLS]}}^{(L)} & \text{DINOv2}
\end{cases}$.

All features are extracted in a single forward pass per image, with gradients disabled, and cached for downstream analysis.

% ======================================================================
\subsection{Spectral Accessibility Analysis}
\label{subsec:analysis}
% ======================================================================

Spatial-frequency information, defined via band energies $E_k(x)$ (Section~3), is analyzed across layer-wise representations by measuring its linear accessibility and isolating changes induced by learned transformations from those explained by dimensionality reduction.

% ----------------------------------------------------------------------
\subsubsection{Accessibility and Spectral Metrics}
\label{subsubsec:metrics}
% ----------------------------------------------------------------------

Accessibility is defined as the degree to which spatial-frequency content can be linearly decoded from a representation \citep{alain2016understanding}. For each frequency band $k$ and representation $\mathbf{z} \in \{\mathbf{z}_H, \mathbf{z}_M, \mathbf{z}_V, \mathbf{z}_P\}$, a separate Ridge regression model is fit on the training set:
\begin{equation}
(\hat{\mathbf{w}}_k, \hat{b}_k) = \arg\min_{\mathbf{w}, b} \frac{1}{|D_{\text{train}}|} \sum_{x \in D_{\text{train}}} \left( E_k(x) - \mathbf{w}^\top \mathbf{z} - b \right)^2 + \lambda \|\mathbf{w}\|_2^2
\end{equation}
with $\lambda = 1.0$, where $\mathbf{z}$ is the representation of input $x$.

Accessibility is evaluated on the test set using Pearson correlation:
\begin{equation}
  \alpha_k(\mathbf{z}) =
  \text{Corr}_{x \sim \mathcal{D}_{\text{test}}}
  \bigl(E_k(\mathbf{x}), \hat{\mathbf{w}}_k^\top \mathbf{z} + \hat{b}_k \bigr).
\end{equation}
This measure is scale-invariant and reflects the linear relationship between predicted and true band energies.

To compare representations, the \emph{Spectral Decay Rate} (SDR) is defined relative to the pre-projection representation $\mathbf{z}_V$:
\begin{equation}
  \text{SDR}_k(\mathbf{z}) =
  \frac{\alpha_k(\mathbf{z}_V) - \alpha_k(\mathbf{z})}
       {\alpha_k(\mathbf{z}_V)},
\end{equation}
where $\mathbf{z} \in \{\mathbf{z}_H, \mathbf{z}_M, \mathbf{z}_P\}$. A positive value indicates reduced accessibility relative to $\mathbf{z}_V$.

To isolate effects beyond dimensionality reduction, we define the \emph{Residual Spectral Loss} (RSL):
\begin{equation}
  \text{RSL}_k(\mathbf{z}) =
  \text{SDR}_k(\mathbf{z}) -
  \overline{\text{SDR}}_k^{\,\text{rand}} ,
\end{equation}
where $\overline{\text{SDR}}_k^{\,\text{rand}}$ denotes the baseline decay induced by random projections (Section~\ref{subsubsec:baseline}). Positive values indicate additional attenuation, while negative values indicate better preservation relative to compression alone.

% ----------------------------------------------------------------------
\subsubsection{Dimensionality-Controlled Baseline}
\label{subsubsec:baseline}

To quantify the effect of dimensionality reduction alone, we construct a random projection baseline following the Johnson--Lindenstrauss framework~\citep{johnson1984extensions,dasgupta2003elementary}. For CLIP, which maps $\mathbf{z}_V \in \mathbb{R}^D$ to $\mathbb{R}^d$, we generate $M=20$ random semi-orthogonal projection matrices.

For each $m = 1, \dots, M$, we sample a Gaussian matrix $G_m \sim \mathcal{N}(0,1)^{D \times d}$ and compute its QR decomposition $G_m = Q_m R_m$, where $Q_m^\top Q_m = I_d$. The projected representation is
\begin{equation}
\mathbf{z}_{\text{rand}}^{(m)} = \mathbf{z}_V Q_m.
\end{equation}
Accessibility and SDR are computed for each projection, and the baseline is obtained by averaging:
\begin{equation}
\overline{\text{SDR}}_k^{\,\text{rand}} = \frac{1}{M} \sum_{m=1}^{M} \text{SDR}_k(\mathbf{z}_{\text{rand}}^{(m)}). \end{equation}
For DINOv2, we instead apply square orthogonal transformations ($d = D$) to provide a baseline that preserves dimensionality while removing learned dependencies between the representation and the spectral targets. Because DINOv2 does not reduce dimensionality ($d = D$), this baseline controls for the reorganization of information rather than compression, isolating effects due to learned transformations. Any non-zero RSL therefore reflects effects beyond this baseline.

% ----------------------------------------------------------------------
\subsubsection{Statistical Inference}
\label{subsubsec:bootstrap}

We estimate uncertainty in RSL due to finite test-set sampling using the nonparametric bootstrap \citep{efron1994introduction}. We generate $B=1000$ bootstrap resamples of the test set (with replacement) and recompute accessibility, SDR, and RSL with fixed probes. This isolates variability due to the test set rather than probe estimation and yields an empirical distribution of $\mathrm{RSL}_k$.

The $95\%$ confidence interval (CI) is computed using the percentile method:
\begin{equation}
\mathrm{CI}_{95\%} =
\bigl[\mathrm{RSL}_k^{(0.025)}, \mathrm{RSL}_k^{(0.975)}\bigr].
\end{equation}
A frequency band exhibits a statistically significant RSL when its confidence interval excludes zero.

\begin{figure}[!t]
  \centering
  % --- Row (a): CLIP ViT-L/14 ---
  \begin{subfigure}[t]{\textwidth}
    \centering
    \begin{minipage}[t]{0.48\textwidth}
      \centering
       \includegraphics[ trim = 0 0 0 36, clip, scale=0.33]{{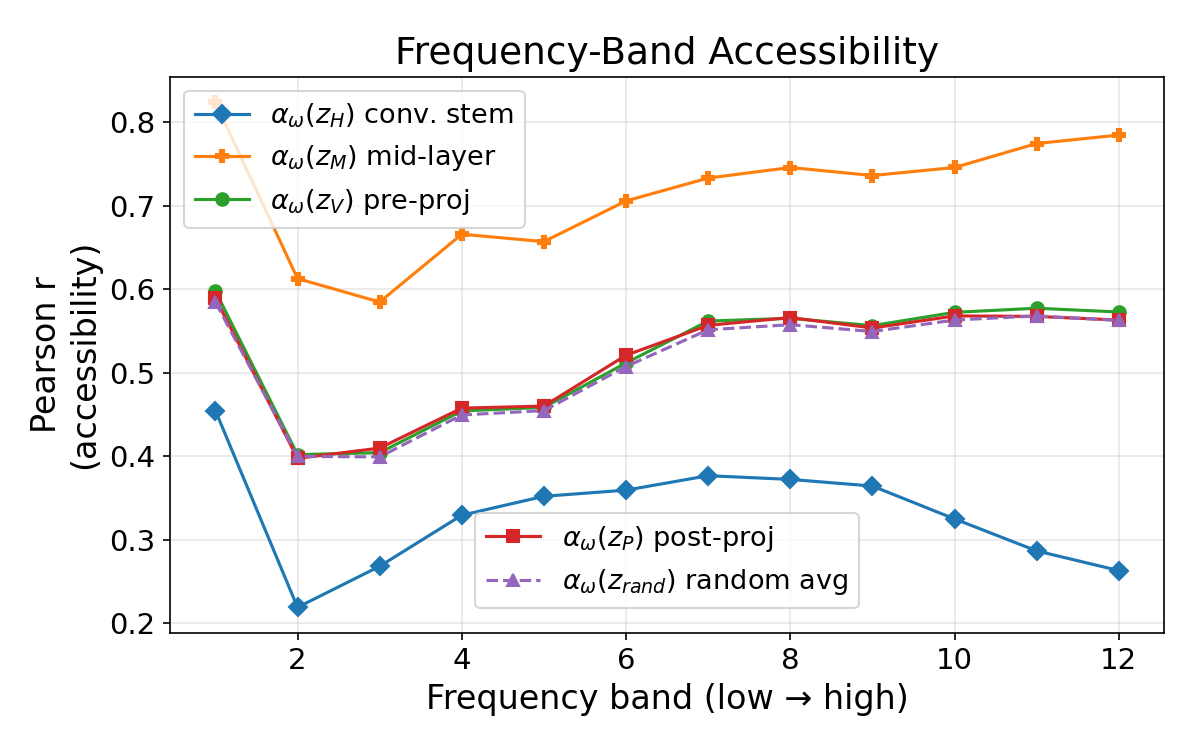}}
    \end{minipage}\hfill
    \begin{minipage}[t]{0.48\textwidth}
      \centering
   \includegraphics[ trim = 0 0 0 36, clip, scale=0.33]{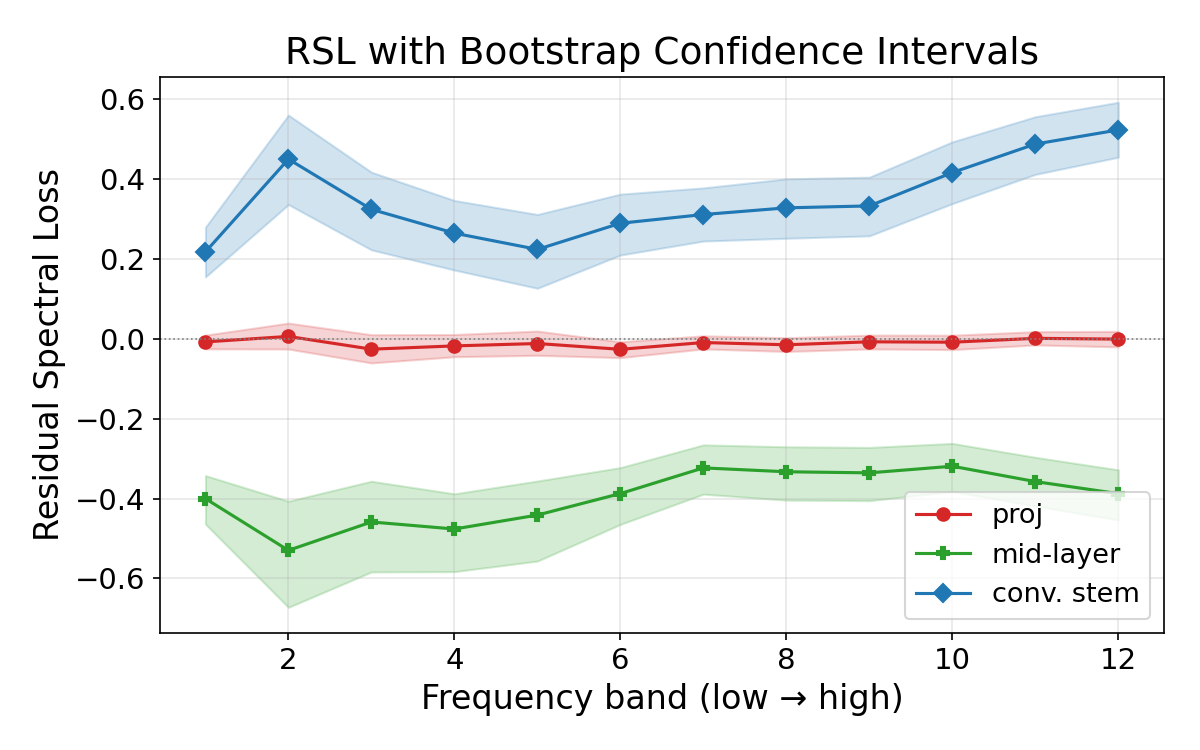}
    \end{minipage}
    \vspace{-4mm}
    \caption{\small CLIP ViT-L/14}
    \label{fig:clip14}
  \end{subfigure}

  \vspace{1em}

\begin{subfigure}[t]{\textwidth}
    \centering
    \begin{minipage}[t]{0.48\textwidth}
      \centering
     \includegraphics[ trim = 0 0 0 36, clip, scale=0.33]{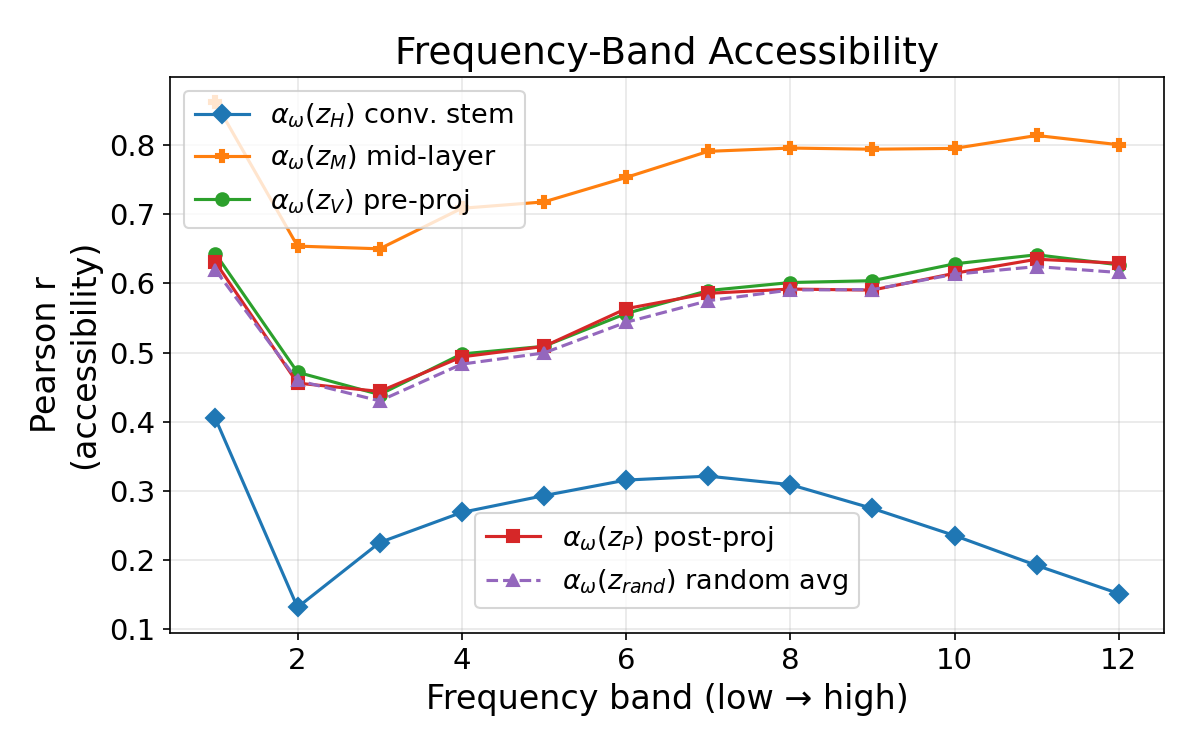}
    \end{minipage}\hfill
    \begin{minipage}[t]{0.48\textwidth}
      \centering
         \includegraphics[ trim = 0 0 0 36, clip, scale=0.33]{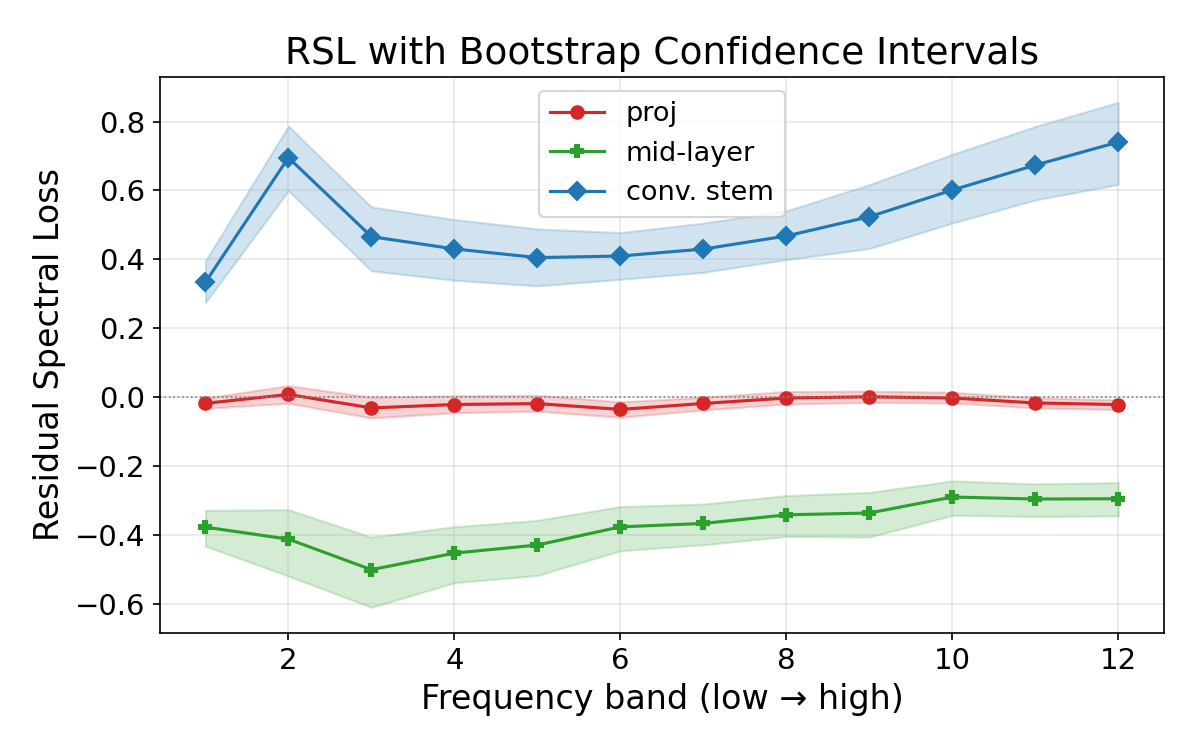}
    \end{minipage}
    
    \vspace{-4mm}
    \caption{\small CLIP ViT-B/32}
    \label{fig:clip32}
  \end{subfigure}

  \vspace{1em}

  \begin{subfigure}[t]{\textwidth}
    \centering
    \begin{minipage}[t]{0.48\textwidth}
      \centering

    \includegraphics[ trim = 0 0 0 36, clip, scale=0.33]{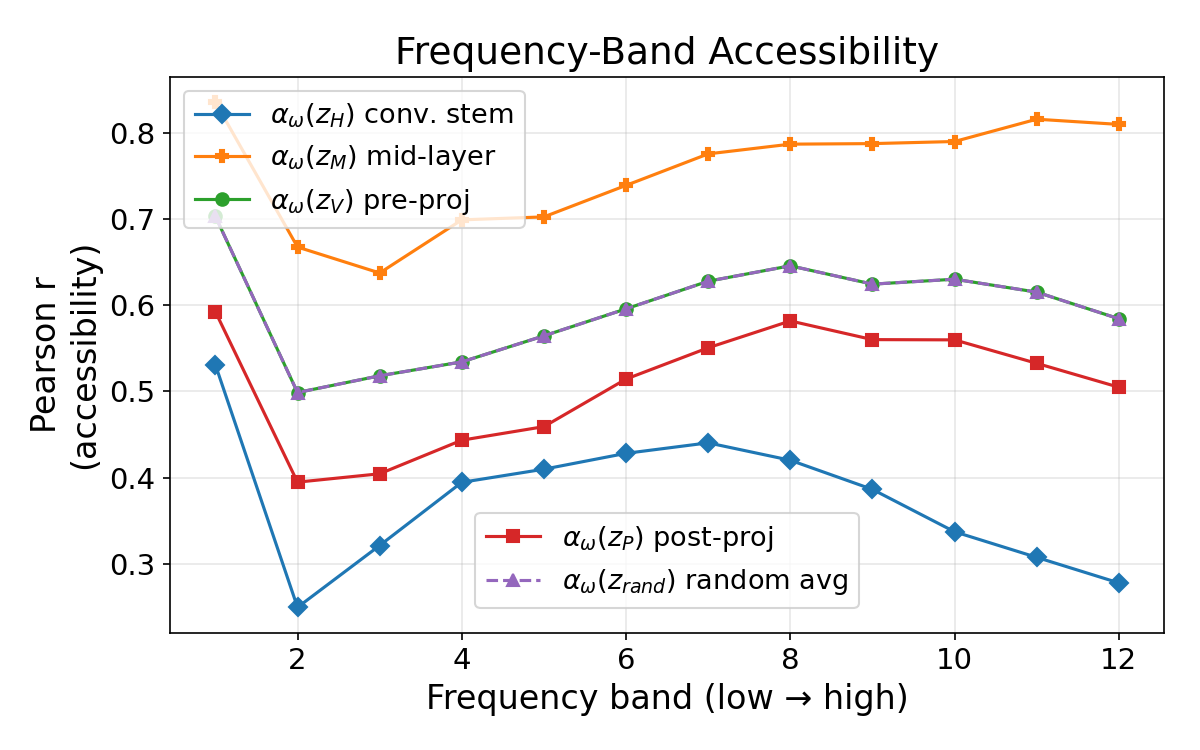}
  
    \end{minipage}\hfill
    \begin{minipage}[t]{0.48\textwidth}
      \centering
   
    \includegraphics[ trim = 0 0 0 36, clip, scale=0.33]{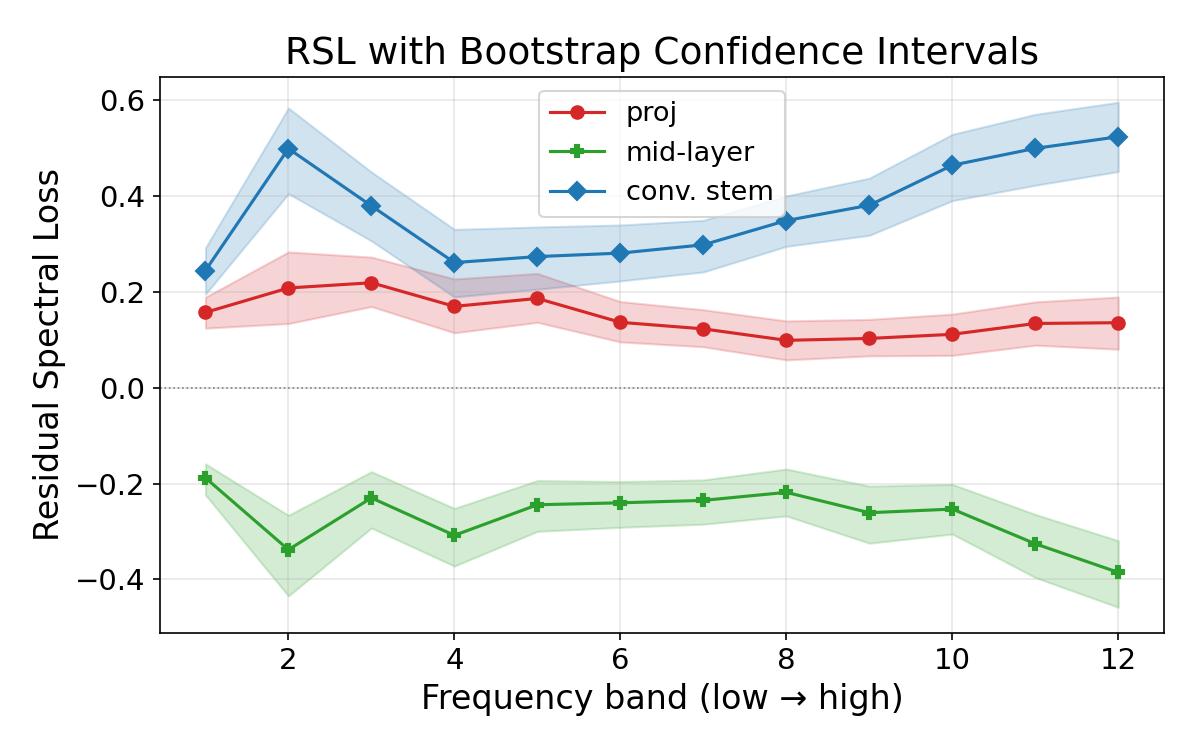}
  
    \end{minipage}
    \vspace{-4mm}
    \caption{\small DINOv2}
    \label{fig:dino}
  \end{subfigure}
  \vspace{-2mm}
  \caption{\small Spatial-frequency accessibility (left) and Residual Spectral Loss (right) on ImageNet.
Left panels plot per-band accessibility $\alpha_k$ across layers; right panels show RSL with 95\% bootstrap confidence intervals. COCO results show the same qualitative patterns (Appendix).}
  % \textbf{Spatial-frequency accessibility (left) and Residual Spectral Loss (right) on ImageNet for all three encoders.} Each row shows one model; left panels plot per-band accessibility $\alpha_k$ at each representational stage, right panels show RSL with 95\% bootstrap CIs. \textbf{(a)}~CLIP ViT-L/14: the mid-layer (orange) dominates across all bands; pre- and post-projection curves nearly overlap, and the projection RSL (red) is indistinguishable from zero. \textbf{(b)}~CLIP ViT-B/32: despite more aggressive compression ($768 {\to} 512$), the projection RSL remains negligible; the conv1 deficit (blue) is the steepest observed, with band-2 RSL reaching $+0.70$. \textbf{(c)}~DINOv2: the output $\mathbf{z}_P$ (\texttt{[CLS]}, red) sits visibly below $\mathbf{z}_V$ (green); the projection RSL is significantly positive at all 12 bands, peaking at low-to-mid frequencies---the \texttt{[CLS]} attention mechanism systematically attenuates spectral content. COCO results are qualitatively identical (Appendix).}
  \label{fig:main_results}
  \vspace{-2mm}
\end{figure}

\section{Results}
\label{sec:results}

We evaluate spectral accessibility across CLIP ViT-L/14, CLIP ViT-B/32, and DINOv2 on ImageNet-1K and COCO. Using RSL to factor out compression effects, we identify consistent frequency-dependent patterns across all configurations. Unless otherwise noted, we report ImageNet results; COCO results are qualitatively similar and are provided in the appendix due to space constraints.

% We evaluate the full analysis pipeline
%  on three frozen vision
% encoders---CLIP ViT-L/14, CLIP ViT-B/32, and DINOv2---using both
% ImageNet-1K and COCO ($N{=}10{,}000$ images each),
% for six configurations in total.  In every case we extract the four
% representational stages $\mathbf{z}_H$, $\mathbf{z}_M$, $\mathbf{z}_V$,
% $\mathbf{z}_P$ (Table~\ref{tab:feature_summary}), compute accessibility via
% Ridge probes with $\lambda{=}1.0$, and derive RSL using $M{=}20$ random
% projections and $B{=}1{,}000$ bootstrap resamples.  Unless otherwise noted, all
% patterns described below hold on both ImageNet and COCO; we report ImageNet
% numbers and note quantitative differences only where they are informative.

\subsection{Spatial-Frequency Accessibility Peaks at Intermediate Depth}
\label{subsec:depth_trajectory}

The accessibility profiles (Figure~\ref{fig:main_results}, left panels) present spatial-frequency accessibility across layers for the three encoders and reveal a consistent non-monotonic trajectory through the network. Across all models and datasets, mean accessibility follows
\begin{equation}
  \bar{\alpha}(\mathbf{z}_H)
  \;<\;
  \bar{\alpha}(\mathbf{z}_P)
  \;<\;
  \bar{\alpha}(\mathbf{z}_V)
  \;<\;
  \bar{\alpha}(\mathbf{z}_M),
  \label{eq:ordering}
\end{equation}
and the per-band ordering $\alpha_k(\mathbf{z}_H) < \alpha_k(\mathbf{z}_P) < \alpha_k(\mathbf{z}_M)$ holds across all frequency bands. Spatial-frequency information therefore does not accumulate monotonically through the encoder. It increases from $\mathbf{z}_H$ to $\mathbf{z}_M$, peaks at intermediate depth, and decreases toward the final representation.

\paragraph{Convolutional stem exhibits limited spectral accessibility.}
At $\mathbf{z}_H$, accessibility is concentrated at low frequencies. Band~1 accessibility is approximately 0.4--0.6, but drops sharply at band~2 by roughly 0.25--0.30 in absolute terms, approximately a halving within a single logarithmic step. Beyond band~2 the curve remains suppressed, with the highest-frequency band recovering only to approximately 0.15--0.35. This pattern reflects the limited receptive field of the convolutional stem: a single $14{\times}14$ (or $32{\times}32$) patch captures coarse spatial structure but does not encode relationships across distant regions. This is consistent with patch embeddings that act as a low-pass filter. Consequently, linearly decodable information is concentrated in low-frequency components.

\paragraph{Intermediate layers increase accessibility, followed by consolidation.}
By the midpoint of the transformer stack ($\ell_{\text{mid}}$), accessibility rises substantially and is no longer confined to low frequencies. This increase is not limited to a specific frequency range: $\alpha_k(\mathbf{z}_M)$ remains high at every band, including those that are poorly represented at $\mathbf{z}_H$. The early transformer blocks therefore integrate cross-patch dependencies into a representation from which spatial-frequency content across the full spectrum becomes linearly decodable. From $\mathbf{z}_M$ to the final pre-projection representation $\mathbf{z}_V$, accessibility decreases moderately. This reduction is approximately uniform across the spectrum. The resulting representation retains substantial spectral information while reflecting a shift toward task-relevant invariances.

\begin{table}[!t]
 \caption{\small Quantitative Summary of Accessibility and RSL Across Models and Datasets}
  \label{tab:results}
  
  \centering
  \small
  %\resizebox{\textwidth}{!}{%
  \begin{threeparttable}
  \renewcommand{\arraystretch}{1}
  \begin{tabular}{
      ll
      @{\hskip 20pt} cccc
      @{\hskip 20pt} rr @{\hskip 20pt} r @{\hskip 20pt} r
    }
    \toprule

    & &
    \multicolumn{4}{c}{\textbf{Accessibility} ($\bar{\alpha}$)}
    & \multicolumn{4}{c}{\textbf{RSL (mean)}} \\

    \cmidrule(lr){3-6} \cmidrule(lr){7-10}

    & &
    $\mathbf{z}_H$ & $\mathbf{z}_M$ & $\mathbf{z}_V$ & $\mathbf{z}_P$
    & \multicolumn{2}{c}{$\mathbf{z}_P$}
    & $\mathbf{z}_M$ & $\mathbf{z}_H$ \\

    \cmidrule(lr){7-8}

    \textbf{Model} & &
    & & &
    & \small\textit{low} & \small\textit{high}
    & & \\

    \midrule

    \multirow{2}{*}{CLIP L/14}
      & IN
      & .331 & .714 & .520 & .518
      & $-.011$ & $\mathbf{-.004}$
      & $\mathbf{-.396}$ & $+.347$ \\

      & CO
      & .371 & .718 & .543 & .540
      & $\mathbf{+.004}$ & $-.016$
      & $-.344$ & $\mathbf{+.297}$ \\

     % \addlinespace[5pt]
      \cmidrule(lr){2-10}

    \multirow{2}{*}{CLIP B/32}
      & IN
      & .260 & .761 & .568 & \textbf{.562}
      & $-.015$ & $-.010$
      & $-.372$ & $+.515$ \\

      & CO
      & .344 & \textbf{.764} & .570 & .556
      & $+.024$ & $-.010$
      & $-.379$ & $+.362$ \\

    % \addlinespace[5pt]
     \cmidrule(lr){2-10}

    \multirow{2}{*}{DINOv2}
      & IN
      & .375 & .754 & .595 & .508
      & $+.189$ & $+.121$
      & $-.269$ & $+.371$ \\

      & CO
      & \textbf{.420} & .753 & \textbf{.626} & .553
      & $+.142$ & $+.105$
      & $-.210$ & $+.327$ \\

    \bottomrule
  \end{tabular}
  \begin{tablenotes}[flushleft]
\small  \item IN = ImageNet, CO = COCO. Accessibility shows mean Pearson $r$ per layer.
RSL reports mean values, with low and high for bands 1--4 and 9--12. Highest $\bar{\alpha}$ and smallest $|\text{RSL}|$ are shown in \textbf{bold}.
\end{tablenotes}
  \vspace{-2mm}

\end{threeparttable}
\end{table}

\paragraph{Quantitative summary.}
Table~\ref{tab:results} provides a quantitative summary of these trends. Mean accessibility ranges from 0.26 to 0.42 at $\mathbf{z}_H$ and from 0.71 to 0.76 at $\mathbf{z}_M$, corresponding to an approximately twofold to threefold increase relative to $\mathbf{z}_H$, depending on the model. The decrease from $\mathbf{z}_M$ to $\mathbf{z}_V$ is more moderate, on the order of 17\%--27\%, and is consistent across configurations.

\subsection{The Output Transformation: Language Projection Versus \texttt{[CLS]} Attention}
\label{subsec:output_transform}

The depth trajectory from $\mathbf{z}_H$ through $\mathbf{z}_V$ is shared across all three architectures. The two model families diverge only at the final step, the transformation from $\mathbf{z}_V$ to the output $\mathbf{z}_P$, and it is precisely this step that the RSL is designed to characterize.

\paragraph{CLIP's learned projection is spectrally neutral.}
For both CLIP variants, the gap between $\mathbf{z}_V$ and $\mathbf{z}_P$ is negligible, with mean differences of 0.002 for ViT-L/14 and 0.006 for ViT-B/32 on ImageNet (Table~\ref{tab:results}). The RSL analysis (Figure~\ref{fig:main_results}, right panels) confirms that this small change is fully explained by dimensionality reduction. Across all CLIP configurations, $\text{RSL}_k(\mathbf{z}_P)$ remains close to zero at every frequency band, never exceeding approximately $\pm 0.035$, with no consistent direction across bands. The same pattern holds for ViT-B/32, whose projection reduces dimensionality more aggressively ($D{=}768 \to d{=}512$). The learned projection $W_{\text{proj}}$ therefore maps $\mathbf{z}_V$ into the shared vision-language space without introducing additional frequency-dependent distortion beyond that expected from compression.

\paragraph{DINOv2's \texttt{[CLS]} attention reduces spectral accessibility.}
DINOv2 exhibits a notably different behavior. The gap between $\mathbf{z}_V$ and $\mathbf{z}_P$ is substantially larger, at 0.087 on ImageNet and 0.073 on COCO (Table~\ref{tab:results}). Because both representations have the same dimensionality ($D{=}768$), this difference reflects a genuine loss in spectral accessibility. The RSL profiles (Figure~\ref{fig:main_results}, right panels) show that $\text{RSL}_k(\mathbf{z}_P)$ is positive and statistically significant across all frequency bands, with values from approximately $+0.10$ to $+0.22$. The \texttt{[CLS]} token's attention pooling thus goes beyond aggregating patch information and systematically reduces the linear decodability of spatial-frequency content across the spectrum.

The spectral shape of the attenuation is also informative. For DINOv2, the RSL peaks at low-to-mid frequencies (around $0.22$) and decreases toward higher bands (around $0.10$), indicating that the \texttt{[CLS]} token disproportionately suppresses coarse-scale structure. By contrast, CLIP's flat RSL profile exhibits no clear frequency-dependent pattern.

\subsection{The RSL Reveals Structured Depth Effects Beyond Raw Accessibility}
\label{subsec:depth_rsl}

While the accessibility curves establish the ordering in Eq.~\ref{eq:ordering}, the RSL provides a finer diagnostic by factoring out the random-projection baseline and revealing how each stage's spectral content differs from $\mathbf{z}_V$ (Figure~\ref{fig:main_results}, right panels).

\paragraph{Convolutional stem: a frequency-graded deficit.}
$\text{RSL}_k(\mathbf{z}_H)$ is positive at every frequency band and exhibits a consistent pattern across all experiments: a pronounced spike at low-to-mid frequencies (around band~2), followed by a gradual increase toward higher frequencies. The spike reaches approximately $+0.45$ to $+0.70$, representing the strongest per-band effect observed in the study. At higher frequencies, $\text{RSL}_{12}$ reaches approximately $2.1$--$3.2$ times the value at band~1, indicating a substantial increase in inaccessibility. The band-2 spike shows a pronounced deficit at low-to-mid frequencies, while the increase toward higher bands suggests reduced accessibility at higher frequencies, which typically correspond to finer spatial scales.

\paragraph{Mid-layer: broadband improvement.}
$\text{RSL}_k(\mathbf{z}_M)$ is negative at all frequency bands in every experiment, with $95\%$ CIs below zero throughout. This indicates that the mid-layer makes spectral content more linearly accessible than $\mathbf{z}_V$ across the spectrum. In the CLIP models, the largest gains occur at low-to-mid frequencies (around band~2), where RSL reaches approximately $-0.49$ to $-0.53$. This pattern mirrors the frequencies where $\mathbf{z}_H$ shows the largest deficit, suggesting that the early transformer layers improve accessibility in frequency ranges that are initially less accessible in $\mathbf{z}_H$. For DINOv2, the gains are more evenly distributed across frequencies, with clear improvements also observed at higher frequencies (e.g., around $-0.39$), where $\mathbf{z}_H$ exhibits larger deficits.

% \subsection{Cross-Dataset Robustness}
% \label{subsec:cross_dataset}
% % ======================================================================

% All qualitative findings replicate on COCO without exception: the four-stage
% ordering (Eq.~\ref{eq:ordering}), CLIP's spectral neutrality, DINOv2's
% \texttt{[CLS]} attenuation, and the conv1/mid-layer RSL profiles all maintain
% their sign, shape, and statistical significance.  The sole systematic
% difference is that $\bar{\alpha}(\mathbf{z}_H)$ is $0.03$--$0.08$ higher on
% COCO, consistent with the greater low-frequency dominance in multi-object
% scenes.  DINOv2's projection RSL is modestly reduced on COCO (mean $0.118$
% vs.\ $0.149$) but remains significant at every band.

\section{Conclusion}
\label{sec:conclusion}

This work examines how modern vision encoders alter spatial-frequency information using a linear accessibility framework and the proposed Residual Spectral Loss (RSL). Spectral accessibility follows a consistent non-monotonic trajectory across depth, peaking at intermediate layers before decreasing toward the output. This indicates that early transformer layers recover spatial-frequency information not accessible from the convolutional stem, while later layers consolidate representations toward task-relevant invariances.
The analysis also reveals a clear architectural distinction. In CLIP, the projection to a shared vision-language space is largely spectrally neutral, with changes explained by dimensionality reduction. In contrast, DINOv2's \texttt{[CLS]} attention pooling induces a structured and statistically significant loss of spectral information, particularly at low-to-mid frequencies. These results suggest that pooling mechanisms may play a more central role than projection layers in determining spectral properties. They also complement standard performance metrics by revealing how models balance information from the input with downstream objectives. This has implications for architectural choices, particularly in aggregation mechanisms and intermediate features.

This study has several limitations. The analysis is observational and does not establish causal effects of specific architectural components. It relies on linear probes and therefore captures linear accessibility rather than the full information content of the representation, so nonlinear structure may differ. The spectral representation is based on grayscale images and radial frequency bands, and does not capture color or directional components. The evaluation is also limited to ViT architectures and natural image datasets, which may restrict generality. Future work can extend this framework by incorporating nonlinear probes, analyzing color and directional frequency components, and introducing controlled interventions on pooling mechanisms or connector designs to establish causal effects.

\bibliographystyle{plainnat}
\bibliography{references}

\newpage
\section*{Appendix}
\begin{figure}[!h]
  \centering
  % --- Row (a): CLIP ViT-L/14 ---
  \begin{subfigure}[t]{\textwidth}
    \centering
    \begin{minipage}[t]{0.48\textwidth}
      \centering
       \includegraphics[ trim = 0 0 0 36, clip, scale=0.33]{{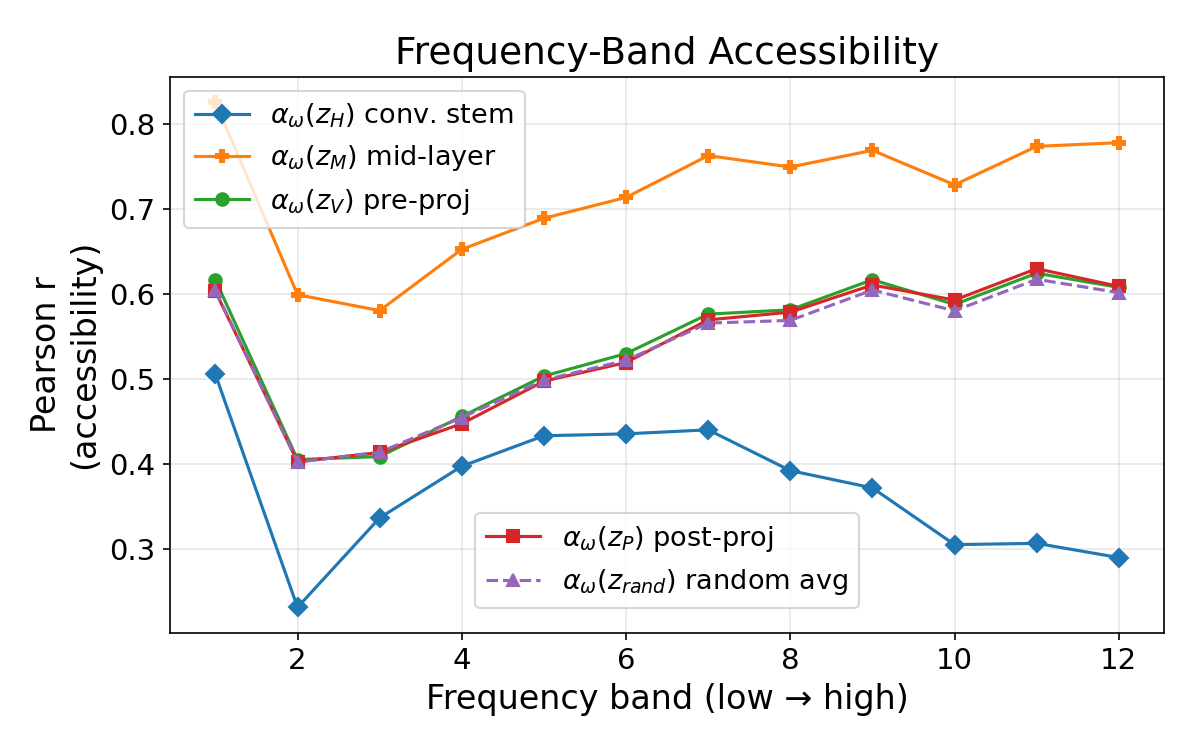}}
    \end{minipage}\hfill
    \begin{minipage}[t]{0.48\textwidth}
      \centering
   \includegraphics[ trim = 0 0 0 36, clip, scale=0.33]{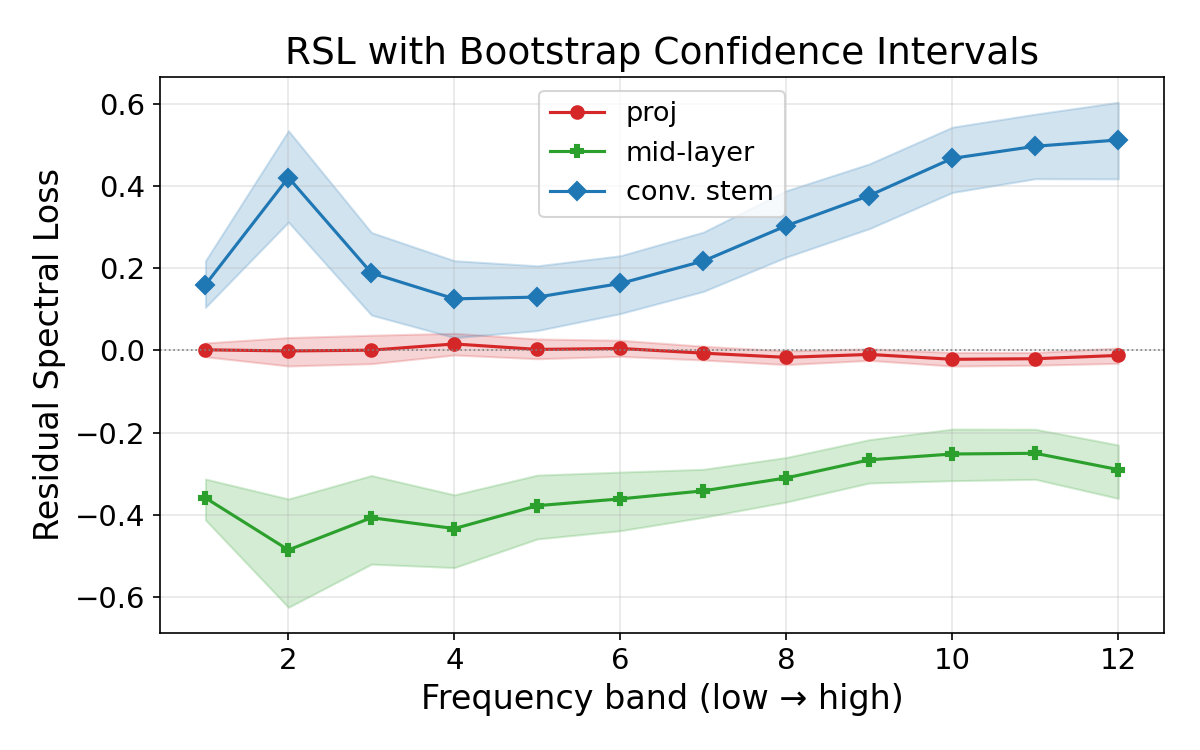}
    \end{minipage}
    \vspace{-2mm}
    \caption{CLIP ViT-L/14}
    \label{fig:clip14}
  \end{subfigure}

  \vspace{0.5em}

\begin{subfigure}[t]{\textwidth}
    \centering
    \begin{minipage}[t]{0.48\textwidth}
      \centering
     \includegraphics[ trim = 0 0 0 36, clip, scale=0.33]{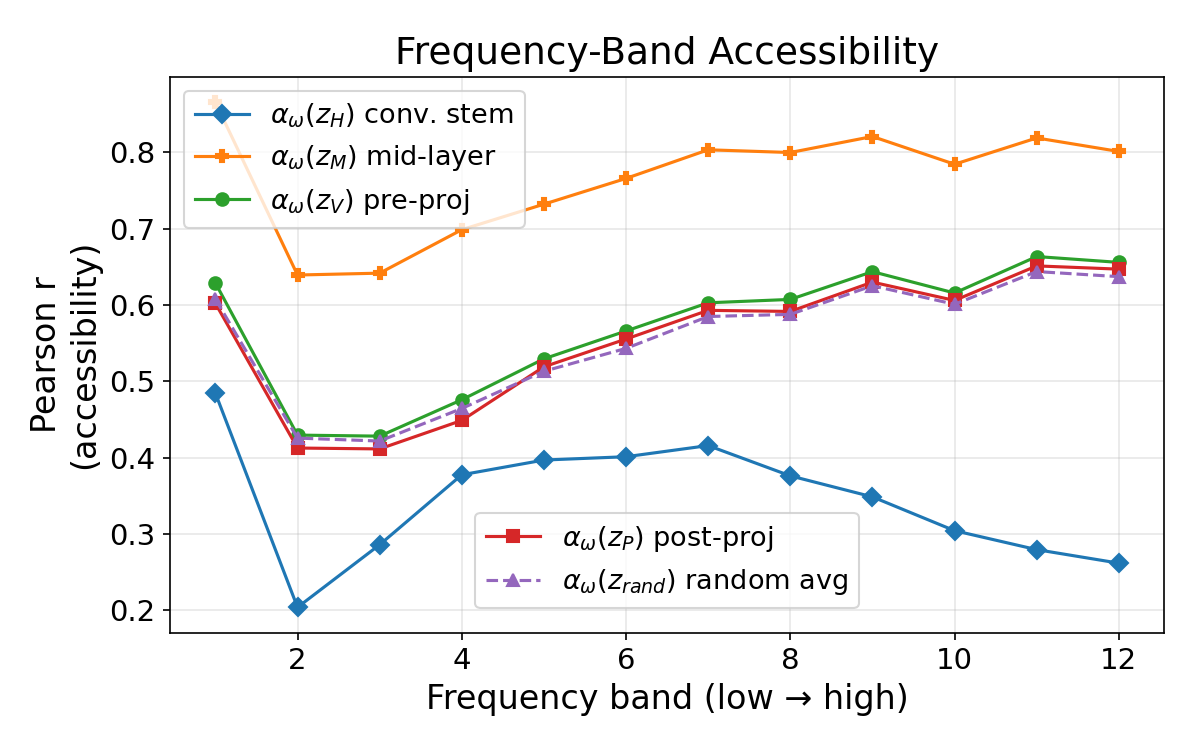}
    \end{minipage}\hfill
    \begin{minipage}[t]{0.48\textwidth}
      \centering
         \includegraphics[ trim = 0 0 0 36, clip, scale=0.33]{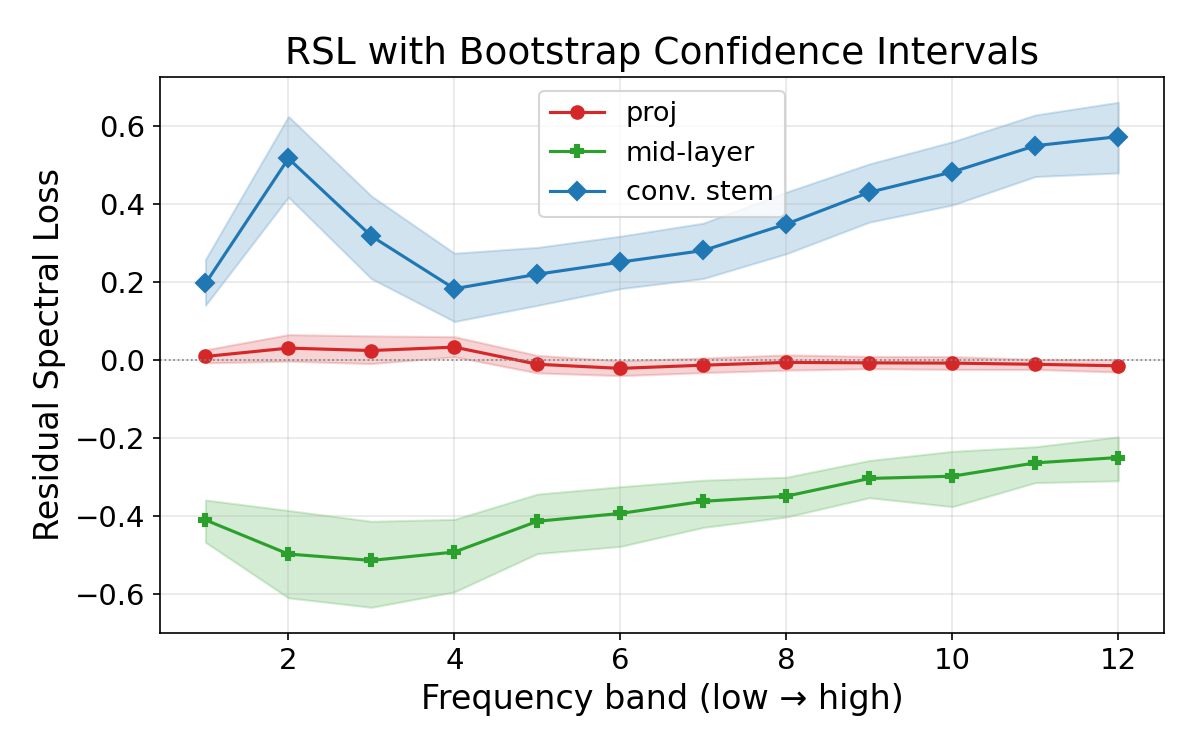}
    \end{minipage}
    
    \vspace{-2mm}
    \caption{CLIP ViT-B/32}
    \label{fig:clip32}
  \end{subfigure}

  \vspace{0.5em}

  \begin{subfigure}[t]{\textwidth}
    \centering
    \begin{minipage}[t]{0.48\textwidth}
      \centering

    \includegraphics[ trim = 0 0 0 36, clip, scale=0.33]{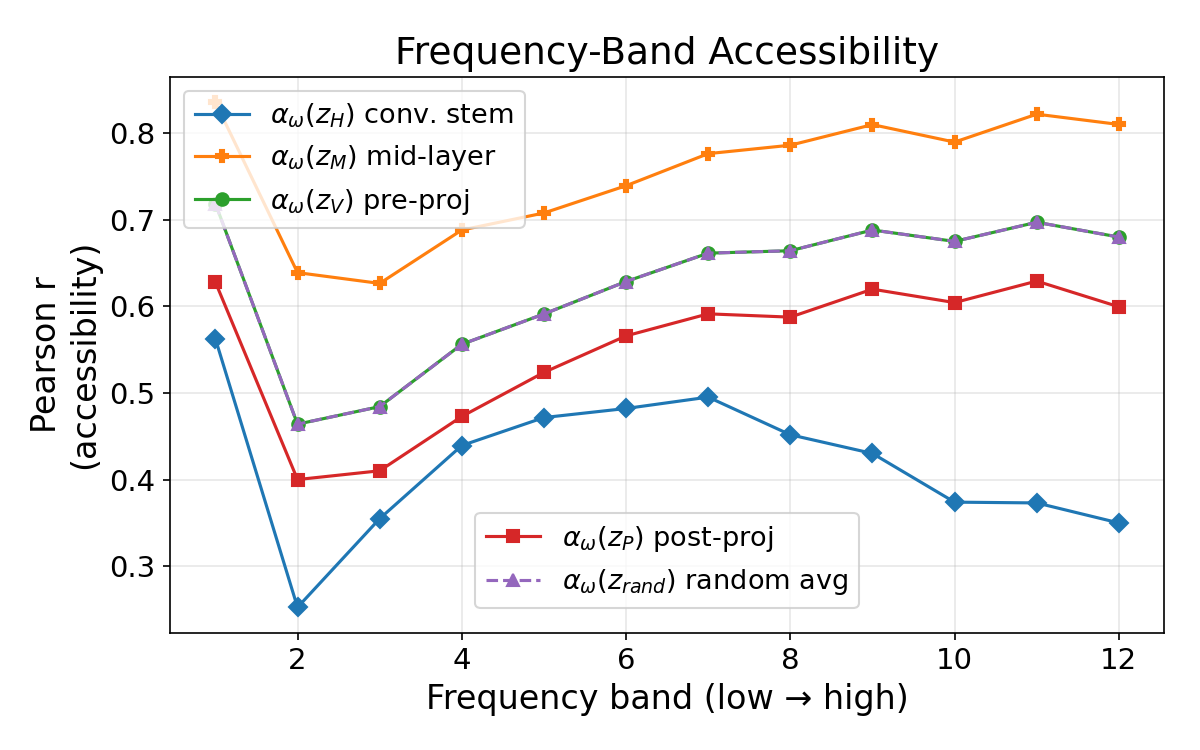}
  
    \end{minipage}\hfill
    \begin{minipage}[t]{0.48\textwidth}
      \centering
   
    \includegraphics[ trim = 0 0 0 36, clip, scale=0.33]{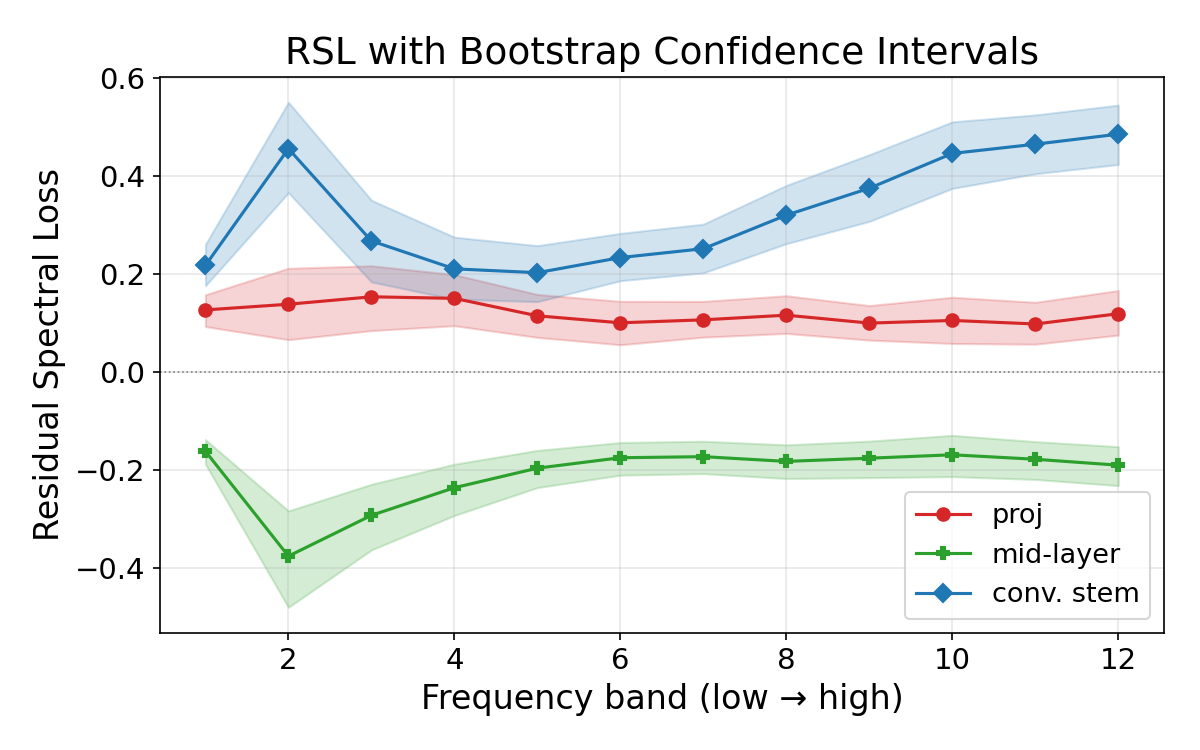}
  
    \end{minipage}
    \vspace{-2mm}
    \caption{DINOv2}
    \label{fig:dino}
  \end{subfigure}
\caption{Spatial-frequency accessibility (left) and Residual Spectral Loss (right) on COCO.
Left panels plot per-band accessibility $\alpha_k$ across layers; right panels show RSL with 95\% bootstrap
confidence intervals.}
\label{fig:appendix_results}
\end{figure}

\end{document}